\pdfoutput=1

\documentclass[11pt]{article}

\usepackage[preprint]{acl}

\usepackage{times}
\usepackage{latexsym}

\usepackage[T1]{fontenc}

\usepackage[utf8]{inputenc}

\usepackage{microtype}

\usepackage{inconsolata}

\usepackage{microtype}
\usepackage{algorithm}
\usepackage{algpseudocode}
\usepackage{amsmath}
\usepackage{amsfonts}
\usepackage{amsthm}
\usepackage{amssymb}

\usepackage{setspace}
\usepackage{graphicx}
\usepackage{caption}
\usepackage{subcaption}
\usepackage{array,multirow}
\usepackage{bm}
\usepackage{adjustbox}
\usepackage{booktabs}
\usepackage{dirtytalk}
\usepackage{stfloats}

\newtheorem{theorem}{Theorem}

\title{DP-MLM: Differentially Private Text Rewriting \\ Using Masked Language Models}

\author{Stephen Meisenbacher, {\bf Maulik Chevli}, {\bf Juraj Vladika}, \and {\bf Florian Matthes} \\
  Technical University of Munich \\
  School of Computation, Information and Technology \\
  Department of Computer Science \\
  Garching, Germany \\
  \texttt{\{stephen.meisenbacher,maulikk.chevli,juraj.vladika,matthes\}@tum.de} \\
}

\begin{document}
\maketitle
\begin{abstract}
The task of text privatization using Differential Privacy has recently taken the form of \textit{text rewriting}, in which an input text is obfuscated via the use of generative (large) language models. While these methods have shown promising results in the ability to preserve privacy, these methods rely on autoregressive models which lack a mechanism to contextualize the private rewriting process. In response to this, we propose \textsc{DP-MLM}, a new method for differentially private text rewriting based on leveraging masked language models (MLMs) to rewrite text in a semantically similar \textit{and} obfuscated manner. We accomplish this with a simple contextualization technique, whereby we rewrite a text one token at a time. We find that utilizing encoder-only MLMs provides better utility preservation at lower $\varepsilon$ levels, as compared to previous methods relying on larger models with a decoder. In addition, MLMs allow for greater customization of the rewriting mechanism, as opposed to generative approaches. We make the code for \textsc{DP-MLM} public and reusable, found at \url{https://github.com/sjmeis/DPMLM}.
\end{abstract}

\section{Introduction}
\begin{figure*}[htbp]
    \centering
    \includegraphics[scale=0.45]{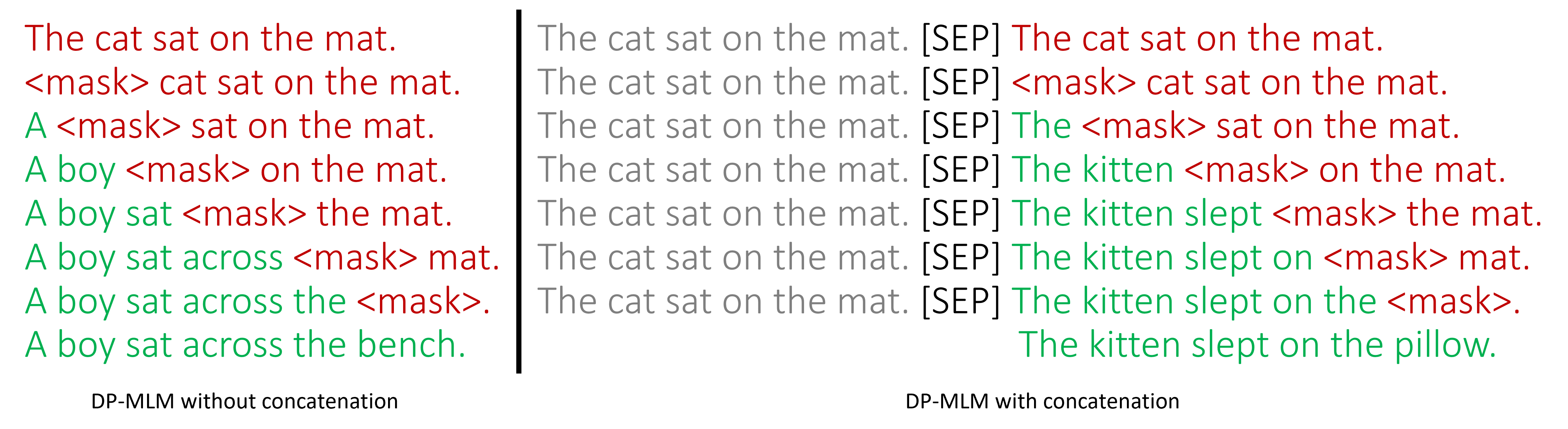}
    \caption{An example of Differentially Private Text Rewriting using Masked Language Models (\textsc{DP-MLM}). The left side shows a real example without contextualization, and the right shows the same example with contextualization. As can be seen, providing a concatenated context sentence (the original sentence) guides the private rewriting process to be more semantically similar than if performed without contextualization.}
    \label{fig:example}
\end{figure*}

The study of Differential Privacy (DP) in NLP investigates the integration of the privacy guarantee offered by DP to the textual domain. This is especially timely as concerns of privacy vulnerabilities in NLP models, particularly LLMs, continue to rise in tandem with recent advances in AI.

Looking to DP for safeguarding privacy in text processing is a promising avenue of research, bringing about novel techniques in recent years ranging from DP optimization techniques \cite{Abadi2016DeepLW}, DP language models \cite{Igamberdiev.2023.ACL}, or DP text privatization methods \cite{feyisetan_balle_2020,mattern-etal-2022-limits, Utpala2023LocallyDP}. DP in NLP does not come without its challenges, however \cite{klymenko-etal-2022-differential, mattern-etal-2022-limits}, among them the balance between privacy and utility, as well as the general reasoning of DP in unstructured domains such as text.

State-of-the-art DP text rewriting methods focus on paraphrasing as a proxy for text privatization, either via directly fine-tuning a paraphrase model \cite{mattern-etal-2022-limits}, or by prompting a LLM to generate a rephrased version of an input text \cite{Utpala2023LocallyDP}. By incorporating a DP mechanism into the generation of output tokens, the generated text aims to privatize the input while still maintaining utility and semantic similarity.

The above works do not consider encoder-only models, such as BERT-based models, and furthermore, the advantages of doing so have not been studied in juxtaposition to models with decoders. Therefore, we are motivated to extend DP text rewriting beyond the usage of autoregressive generative models, guided by the following question:

\begin{center}
    \textit{How can one leverage Masked Language Models (MLMs) to achieve contextually relevant, yet privacy-preserving text rewriting?}
\end{center}

To answer this question, we propose \textsc{DP-MLM}, a differentially private text rewriting mechanism leveraging masked token prediction in BERT-based models. We design a rewriting mechanism that incorporates the contextual information from an input sentence to produce a rewritten output that is both utility- and privacy-preserving. This customization is enabled by building a rewriting mechanism on top of an MLM, as opposed to relying on a decoder to generate privatized outputs. To test utility and privacy, we conduct empirical experiments that validate the ability of \textsc{DP-MLM} to rewrite texts in a way that preserves meaning yet still empirically defends against adversarial privacy attacks.

Our work makes the following contributions to the study of differentially private text rewriting:
\begin{enumerate}
    \itemsep 0em
    \item To the best of our knowledge, we are the first work to propose the use of BERT-based models for DP text rewriting.
    \item We design a rewriting mechanism, \textsc{DP-MLM}, which leverages a contextualization technique not before used for text privatization.
    \item With \textsc{DP-MLM}, we surpass previous SOTA methods for private text rewriting on many benchmarks in both utility and privacy, particularly at lower per-token $\varepsilon$ budgets.
\end{enumerate}

These contributions support our hypothesis that MLMs are effective tools for utility-preserving text privatization, and its observed strengths are analyzed at the conclusion of this work.

\section{Related Work}
Natural language contains sensitive information \cite{Brown2022WhatDI, McMahan2017LearningDP}. DP enables the training of Machine Learning (ML) models on sensitive texts with a guarantee that the model will not leak more sensitive data than a pre-defined value \cite{Pan2020, McMahan2017LearningDP, carlini2021extracting}. In the field of NLP, there are two primary notions of integrating DP for downstream applications. The first approach is to collect user texts at a central location and train a model on the texts using a differentially-private optimization technique like DP-SGD \cite{Abadi2016DeepLW} \cite{ponomareva-etal-2022-training, McMahan2017LearningDP, Kerrigan2020DifferentiallyPL}. This approach is known as \textit{global} or \textit{central} DP. In contrast, the second approach is to apply a DP mechanism on texts \textit{locally}, i.e., on the user side, before sharing the privatized texts with a central aggregator. This notion is called Local Differential Privacy (LDP). LDP is a stricter notion of DP as compared to the formerly defined central DP \cite{feyisetan_balle_2020}.

The earliest application of LDP to the task of text privatization considers a sentence as independent sequences of words \cite{fernandes2019generalised, feyisetan_balle_2020}. As a result, the privatized sentence is generated by perturbing a sentence word-by-word, normally by introducing calibrated noise to word embeddings \cite{yue, Chen2022ACT, carvalho2023tem, meisenbacher20241}. These methods do not consider the grammatical and contextual information while generating a private sentence \cite{mattern-etal-2022-limits}. Moreover, they utilize the generalized notion of metric DP, which increases the utility of the generated text but makes comparative evaluations challenging.

Other LDP methods operate at higher levels in the syntactic hierarchy, such as the sentence-level, and directly generate a privatized  text, either by adding noise to the latent representations of the text \cite{Igamberdiev.2023.ACL} or using additional public information \cite{Meehan2022SentencelevelPF}. By construction, these methods take into account the grammar and context of the input text while privatizing it. These methods provide stricter privacy guarantees as compared to the aforementioned word-level mechanisms but come at a significant utility cost \cite{Igamberdiev.2023.ACL}.

Recent works for LDP in NLP take a different approach and leverage Language Models to generate privatized text \cite{mattern-etal-2022-limits, Utpala2023LocallyDP}. They model the task of text privatization as a text paraphrasing task, or more generally, text rewriting. In leveraging such models to rewrite text, generation is performed by tokens being sampled in a fashion that satisfies local DP. We build upon the foundations of these methods, and aim to improve upon the utility- and privacy-preserving capabilities of such an approach.

\section{Foundations}
\subsection{Masked Language Modeling}
A Masked Language Model (MLM), such as BERT \cite{devlin-etal-2019-bert}, includes as one of its two pre-training objectives the task of \textit{masked token prediction}. Here, the model is trained to predict a randomly masked token in a sentence by conditioning its probability not only on the tokens that are to its left in the sentence but also to its right. Thus, the masked token is filled by the most suitable word deemed by the MLM, which is achieved by considering the \emph{complete} context of the sentence.

If the $l^{\text{th}}$ token of a sentence $s$ containing $n$ tokens in sequence $w_1 \cdot w_2 \cdots w_n$ is masked, the probability of the masked token being a word $v$ from the vocabulary $\mathcal{V}$, as modeled by an MLM, is
\begin{displaymath}
    Pr[w_l = v] = Pr[v | w_1\cdot w_2 \cdots w_{l-1} \cdot w_{l + 1} \cdots w_n]  
\end{displaymath}

Although the masked token of a MLM is primarily used in its pre-training task, one can also use the token to replace a given target word in a text. This fact is leveraged by the related tasks of MLM-based Lexical Substitution \cite{zhou-etal-2019-bert} or Lexical Simplification \cite{qiang2020lexical}.


\subsection{Differential Privacy}
Differential Privacy (DP) \cite{dwork2006differential} is a mathematically grounded notion of privacy that provides information-theoretic privacy guarantees while performing computation over a dataset. Given $\varepsilon \geq 0$ and finite sets $\mathcal{W}$ and $\mathcal{V}$, a randomized mechanism $M: \mathcal{W} \to \mathcal{V}$ is an $\varepsilon$-DP mechanism if $\forall c, c^\prime \in \mathcal{C}$ and $\forall v \in \mathcal{V}$, the following condition holds:
\begin{displaymath}
    \frac{Pr[M(c) = v]}{Pr[M(c^\prime) = v]} 
    \leq e^{\varepsilon}
\end{displaymath}

Here $c$ and $c^\prime$ are called \textit{neighboring} or \textit{adjacent} databases. Depending on the notion of adjacency, the unit which is protected by LDP is defined. In our case, any two context sentences $c$ and $c^\prime$ are adjacent. This is expounded upon in the next section.

\subsubsection{Temperature Sampling as an Exponential Mechanism}
\label{sec:temp}
Suppose there is a dataset $D \in \mathcal{X}^n$ and our aim is to derive a value for this dataset from a set of fixed value choices $\mathcal{V} \in \mathcal{Y}$.
Exponential Mechanism can be used here to select, in a private manner, the best choice for a dataset from a set of choices $\mathcal{V}$, with its goodness being determined by a scoring function. The scoring (utility) function $u: \mathcal{X}^n \times \mathcal{Y} \to \mathbb{R}$ maps database and choice pairs, $(D, w)$ with $D\in \mathcal{X}^n$ and $w \in \mathcal{V}$, to possibility scores. The $l_2$-sensitivity of such scoring function is given as
$$\Delta u = \underset{w \in \mathcal{V}}{\max} \underset{D, D^\prime \in \mathcal{X}^n}{\max} |u(D, w) - u(D^\prime, w)|$$

If the choice for $D$ is selected according to probability proportional to $\exp(\frac{\varepsilon u(D, w)}{2 \Delta u})$, then the selection algorithm satisfies DP and this DP mechanism is termed as an Exponential Mechanism \cite{exponential-mechanism}. 

Mapping it to our use case, suppose we have a context $s = w_1 \cdot w_2 \cdot w_3 \cdots w_n$ or $s = w_1 \cdot w_2 \cdots w_{l-1} \cdot w_{l+1} \cdots w_n $, and we want to choose the best token for the masked word $w_l$ from the set of vocabulary. This selection can be done privately through an Exponential Mechanism with scoring function $u: \mathcal{V}^* \times \mathcal{V} \to \mathbb{R}$ that takes as input the whole context $s$ and a word $w$ from the vocabulary set $\mathcal{V}$ and outputs a score for $w$ conditioned on the context $s$. This scoring function is precisely a (Masked) Language Model in our case: it takes the context as input and outputs logits for every word in the vocabulary. To bound the sensitivity of the scoring function, which is necessary for DP, the logit values can be clipped to a predefined range. 

If the logits generated by the MLM are bounded, then the Exponential Mechanism is realized by default if we use temperature sampling to predict the masked token \cite{mattern-etal-2022-limits}. To compare it, if the logit value produced for a word $w \in \mathcal{V}$ with a context $s$ is $u$ and the sampling temperature is set to $T$, the predicted token being $w$ has the probability proportional to $\exp(u / T)$. Comparing it to the exponential mechanism, we can derive that $\varepsilon = \frac{2 \Delta u}{T}$. A detailed DP proof is provided in Section \ref{sec:guarantee}.

\section{Method}
In this section, we outline the design of \textsc{DP-MLM}, particularly its underlying DP mechanism, as well as the text rewriting mechanism.

\subsection{DP Masked Token Prediction}
Given an input sentence $s$ with $n$ tokens, our goal is to privatize each token in the sentence, one token at a time. To privatize a single token $w_l$, we input the entire sentence $s = w_1 \cdots w_{l-1} \cdot w_l \cdot w_{l+1} \cdots w_n$ to an MLM, with $w_l$ replaced by the model's mask token, usually \texttt{<mask>}. Next, we capture the logit values for the masked token index $l$, clip them (described in Section \ref{sec:clip}), and apply Temperature Sampling mechanism, that is equivalent to the Exponential Mechanism as described in Section \ref{sec:temp}. All clipped logit values are first divided by the temperature $T$, which is calculated according to $T = \frac{2 \Delta u}{\varepsilon}$. The resulting values are fed through a softmax function and a token is sampled according to these probabilities. The sampled token $w_p$ then serves as the differentially private token replacement for $w_l$. A proof that this mechanism is DP is found in Section \ref{sec:guarantee}.

\subsection{Rewriting Mechanism}
To rewrite an entire text with the underlying mechanism described above, we design a rewriting mechanism \textsc{DP-MLM} that is outlined in Algorithms \ref{alg:dpmlm} and \ref{alg:rewrite}. Note that for the entirety of this work, we use the \textsc{roberta-base} MLM as our base model.

\begin{algorithm}[htbp]
\caption{\newline DP-MLM Token Replacement}
\label{alg:dpmlm}
    \begin{algorithmic}
        \small
        \Require MLM $M$, \\ context $\texttt{tokens}$, private $\texttt{p\_tokens}$, position $idx$ \\ epsilon $\varepsilon$, \\ clipping values $C = (C_{min}, C_{max})$
        \Ensure Output $\texttt{private\_token}$ at position $idx$
        \\
        \State $T \gets 2\cdot(\vert C_{max} - C_{min}\vert)/ \varepsilon$
        \State $\texttt{p\_tokens}[idx] \gets \texttt{<mask>}$
        \State $\texttt{masked\_sent} \gets concat(\texttt{tokens}, \texttt{[SEP]}, \texttt{p\_tokens})$
        \State $\texttt{logits} \gets M(\texttt{masked\_sent})$
        \State $\texttt{logits} \gets clip\_and\_temp(\texttt{logits}, C, T)$
        \State $\texttt{prob} \gets softmax(\texttt{logits})$
        \State $\texttt{private\_token} \gets sample(\texttt{prob})$
        \State \Return $\texttt{private\_token}$
    \end{algorithmic}
\end{algorithm}

\begin{algorithm}[htbp]
\caption{\newline Text Rewriting using DP-MLM}
\label{alg:rewrite}
    \begin{algorithmic}
        \small
        \Require input sentence $s  = w_1 \cdot w_2 \cdots w_n$, \\ per-token epsilon $\varepsilon$, \\ clipping values $C = (C_{min}, C_{max})$
        \Ensure rewritten (privatized) sentence using DP-MLM
        \\
        \State $\texttt{tokens} \gets tokenize(s)$ 
        \State $\texttt{private} \gets tokens$ 
        \For {$i \in 1...n$}
            \State $\texttt{p} \gets DPMLM(\texttt{tokens}, \texttt{private}, i, \varepsilon, C)$ \Comment{Alg. \ref{alg:dpmlm}}
            \State $\texttt{private}[i] \gets p$
        \EndFor
        \State \Return $detokenize(\texttt{private})$
    \end{algorithmic}
\end{algorithm}

\noindent As described in Algorithm \ref{alg:dpmlm}, contextualization of the \textsc{DP-MLM} mechanism is achieved via the concatenation of the original input sentence, which is given along with the masked sentence as input to the MLM. This simple trick is motivated by a similar approach followed by \cite{qiang2020lexical} for Lexical Simplification. A more intuitive illustrative example of the \textsc{DP-MLM} rewriting mechanism is found in Figure \ref{fig:example}. Note that in our implementation, we do not replace English stopwords, but we leave this as a parameter in our open-source code.

With Algorithm \ref{alg:dpmlm}, we replace a single token from an input text in a DP manner. Thus, the output of one \textsc{DP-MLM} usage is a \textit{privatized} token that is contextually relevant to the text. By using \textsc{DP-MLM} for each token in the input sentence (Algorithm \ref{alg:rewrite}), we design a text rewriting mechanism that leverages the compositionality of DP to output a privately rewritten text with a DP guarantee of $\varepsilon \times len(\texttt{tokens})$. This is formalized for the token and text level in the following.

\subsection{Privacy Guarantees}
\label{sec:guarantee}
Our mechanism $\mathcal{M}$ introduced in Algorithm \ref{alg:dpmlm} satisfies local differential privacy (LDP). For any two adjacent context sentences, $\mathcal{M}$ yields \say{similar} tokens to fill a given masked token with LDP guarantees. Hence, given a predicted masked token, an adversary who only sees the predicted token cannot differentiate with high certainty if the predicted token was due to a context sentence $c$ or $c^\prime$. This results in plausible deniability about the source of the predicted masked token.

Suppose a sentence $s$ consists of $n$ tokens, i.e., $s = w_1 \cdot w_2 \cdots w_n$, and the (masked) token that we want to predict lies at $l^{\text{th}}$ position in the sentence. An MLM models the likelihood of masked token $w_l$ being $v_i \in \mathcal{V}$ as follows:

\begin{align}
    Pr[w_l = v_i] 
    &= Pr[v_i | w_1\cdots w_{l-1} \cdot w_{l + 1} \cdots w_n] \nonumber \\
    & = Pr[v_i | C_l] \nonumber
\end{align}

As stated earlier, we use clipped logits $u$ and temperature sampling to sample the most likely token for the masked token. Hence, for our proposed mechanism $M: \mathcal{V}^+ \to \mathcal{V}$ that takes the context sentence $C_l = w_1\cdot w_2 \cdots w_{l-1} \cdot w_{l + 1} \cdots w_n$ as input and returns a privately selected predicted masked word, its output probability distribution is given by the following equation:
\begin{equation}
    Pr[M(C_l) = v_i] = 
    \frac{ \exp(\frac{u(C_l, v_i)}{T}) }
         { \sum_{j=1}^{|\mathcal{V}|} \exp(\frac{u(C_l, v_j)}{T})}
    \label{eq:mech_prob}
\end{equation}

\begin{theorem}
The proposed mechanism $M$ defined in the equation \ref{eq:mech_prob} satisfies $\varepsilon$-LDP.
    \begin{proof}
    Let $s, s^\prime$ $\in \mathcal{V}^n$. Suppose the $l^{\text{th}}$ token of $s, s^\prime$ is masked and we use our mechanism $M$ to predict a token. The context to be set as input for the Masked Language Model becomes:
    $$C_l := w_1\cdot w_2 \cdots w_{l-1} \cdot w_{l + 1} \cdots w_n$$
    $$C_l^{\prime} := w_1^{\prime}\cdot w_2^{\prime} \cdots w_{l-1}^{\prime} \cdot w_{l + 1}^{\prime} \cdots w_n^{\prime}$$

    The ratio of the probability distribution of application of $M$ on $C_l$ and $C_l^{\prime}$ can be given as:

    \tiny
    \begin{align*}
    \frac{Pr[M(C_l) = v_i]}
         {Pr[M(C_l^{\prime}) = v_i]} 
    &= \frac{ \exp(\frac{u(C_l, v_i)}{T}) }
            { \sum_{j=1}^{|\mathcal{V}|} \exp(\frac{u(C_l, v_j)}{T}) }
       \frac{ \sum_{j=1}^{|\mathcal{V}|} \exp(\frac{u(C_l^{\prime}), v_j}{T})}
            { \exp(\frac{u(C_l^{\prime}), v_i)}{T}) }  \\
    &= \frac{ \exp(\frac{u(C_l, v_i)}{T}) }
            { \exp(\frac{u(C_l^{\prime}, v_i))}{T}) }
        \frac{ \sum_{j=1}^{|\mathcal{V}|} \exp(\frac{u(C_l^{\prime}, v_j)}{T})}
             { \sum_{j=1}^{|\mathcal{V}|} \exp(\frac{u(C_l, v_j)}{T}) }
    \end{align*}

    \small
    \text{\normalsize Solving the first fraction, we get}
    \begin{align*}
    \frac{ \exp(\frac{u(C_l, v_i)}{T}) }
            { \exp(\frac{u(C_l^{\prime}, v_i))}{T}) }
    &= \exp\left( \frac{u(C_l, v_i) - u(C_l^{\prime}, v_i)}{T} \right) \\
    & \leq \exp\left( \frac{\Delta u }{T} \right)
    \end{align*}

    \text{\normalsize Similarly, solving the second fraction, we get}
    \begin{align*}
    \frac{Pr[M(C_l) = v_i]}
         {Pr[M(C_l^{\prime}) = v_i]} 
    & \leq \exp\left( \frac{\Delta u }{T} \right) \exp\left( \frac{\Delta u }{T} \right) \\
    & = \exp\left( 2 \frac{\Delta u }{T} \right) \\
    & = \exp\left( \varepsilon \right)
    \end{align*}
    \end{proof}
\end{theorem}
Hence, $\varepsilon$ can be calculated from the sensitivity of $\Delta u$ and the sampling temperature $T$ as 
$
\varepsilon = 2 \frac{\Delta u }{T}
$

\paragraph{Extending guarantees to a sentence}
The privacy budget required for generating a single masked token using the mechanism $M$ is equal to $\varepsilon$. The privatized sentence of the input sentence is generated by sequentially generating privatized token for each token present in the input sentence. Thus, for rewriting a sentence (or more generally, a text) of $n$ tokens, we are required to call the mechanism $n$ times. Hence, by sequential composition, the total privacy budget spent for rewriting the entire text would be $n\varepsilon$-DP.

\section{Experimental Setup}
In order to evaluate the performance of our proposed method, we design two overarching experiments: (1) utility experiments, and (2) empirical privacy experiments. These are outlined below.

\subsection{Utility Experiments}
Our utility experiments consist of two phases: (1) utility benchmarking of \textsc{DP-MLM}, and (2) comparative utility testing with two other state-of-the-art DP rewriting mechanisms.

\subsubsection{Utility Benchmarking}
\label{sec:benchmarking}
To test the utility-preserving capability of \textsc{DP-MLM}, we measure the utility of rewritten data across a range of $\varepsilon$ values. For this, we utilize the \textsc{GLUE} benchmark \cite{wang2019glue}, which contains 9 separate tasks spanning classification, textual similarity, and textual entailment. For a given $\epsilon$ value, a perturbed dataset (train and validation split) is measured against the non-privatized baseline, in order to measure how well the privatization keeps utility intact.

\paragraph{Model}
For both the non-privatized baseline and all privatized (rewritten) datasets, we fine-tune a \textsc{deberta-v3-base} model \cite{he2021debertav3} for one epoch. The trained model is then evaluated on the validation set (non-privatized or privatized, respectively). All models in this work are trained on a single RTX A6000 GPU, using the Adam optimizer \cite{kingma2017adam} and all default hyperparameters of the Hugging Face Transformers Trainer API. To account for variations in training, all utility results represent the average of 3 runs.

\paragraph{Clipping Values}
\label{sec:clip}
To ensure that the sensitivity of the logit values of our underlying MLM (\textsc{roberta-base}) is bounded, we pre-define the clip value based upon an empirical estimation of the logit value range. Concretely, we measure all logits values from inputting 1000 random text examples from the \textsc{SST2} dataset of \textsc{GLUE}, calculate the mean $\mu$ and standard deviation $\sigma$, and define the clipping values as
$
    (C_{min}, C_{max}) = (\mu, \mu + 4\sigma)
$.
The choice of such a range provides the benefit of a bound sensitivity, while still preserving higher logit values and \say{clamping} lower values.

\paragraph{Epsilon} 
In order to conduct comprehensive tests on varying privacy budgets, we choose the following set of $\varepsilon$ values: $\varepsilon \in \{10, 25, 50, 100, 250\}$.

\begin{table*}[ht!]
\resizebox{\textwidth}{!}{
\begin{tabular}{cl|ll|lll|llll}
\multicolumn{2}{r|}{Task} & \textsc{CoLA} & \textsc{SST2} & \textsc{QQP} & \textsc{MRPC} & \textsc{STSB} & \textsc{MNLI} & \textsc{QNLI} & \textsc{WNLI} & \textsc{RTE} \\ \hline
\multicolumn{2}{r|}{Baseline} &  $84.72_{0.35}$ & $95.72_{0.22}$ & $89.26_{0.05}$ & $84.07_{0.40}$ & $84.57_{0.78}$ & $88.75_{0.03}$ & $93.51_{0.12}$ & $56.34_{0.00}$ & $54.99_{2.98}$  \\ \hline
\multirow{5}{*}{$\varepsilon$} & 10 &  $69.13_{0.00}$ & $68.50_{0.64}$ & $71.86_{0.26}$ & $71.32_{2.31}$ & $6.13_{0.90}$ & $52.86_{0.71}$ & $66.25_{1.11}$ & $53.05_{4.65}$ & $51.50_{1.70}$  \\
 & 25 & $69.77_{0.12}$ & $76.49_{0.94}$ & $74.17_{0.17}$ & $70.67_{1.80}$ & $12.42_{2.45}$ & $56.15_{2.95}$ & $68.54_{5.63}$ & $55.40_{1.33}$ & $51.87_{1.19}$
 \\
 & 50 &  $70.85_{0.86}$ & $84.10_{0.42}$ & $80.53_{0.19}$ & $75.25_{1.31}$ & $26.21_{11.60}$ & $66.99_{0.18}$ & $82.01_{0.03}$ & $52.58_{5.31}$ & $52.47_{0.34}$  \\
 & 100 & $70.05_{0.86}$ & $86.16_{0.39}$ & $82.17_{0.26}$ & $74.35_{1.63}$ & $34.68_{4.20}$ & $69.57_{0.11}$ & $83.56_{0.07}$ & $48.36_{5.67}$ & $53.31_{0.61}$  \\
 & 250 &  $70.18_{0.75}$ & $86.05_{0.55}$ & $82.44_{0.03}$ & $76.39_{0.92}$ & $60.77_{1.49}$ & $70.96_{0.19}$ & $84.68_{0.20}$ & $51.64_{5.67}$ & $51.38_{1.87}$
\end{tabular}
}
\caption{Utility Benchmark Scores for \textsc{DP-MLM}. All scores represent accuracy scores, except for \textsc{STSB}, which is represented by the Pearson-Spearman Correlation score. The metrics are an average of three training runs, and the standard deviation is presented as a subscript. In all cases, a higher score is better.}
\label{tab:utility}
\end{table*}

\begin{table*}[htbp]
\resizebox{\textwidth}{!}{
\begin{tabular}{cc|l|lll|lll|lll}
\multirow{2}{*}{Task} & \multicolumn{1}{l|}{\multirow{2}{*}{$\varepsilon$}} & \multirow{2}{*}{Baseline} & \multicolumn{3}{c|}{DP-MLM} & \multicolumn{3}{c|}{DP-Paraphrase} & \multicolumn{3}{c}{DP-Prompt} \\
 & \multicolumn{1}{l|}{} &  & \multicolumn{1}{c}{BLEU} & \multicolumn{1}{c}{CS} & Acc. & \multicolumn{1}{c}{BLEU} & \multicolumn{1}{c}{CS} & \multicolumn{1}{l|}{Acc.} & \multicolumn{1}{c}{BLEU} & \multicolumn{1}{c}{CS} & Acc. \\ \hline
\multirow{5}{*}{\textsc{CoLA}} & 10 & \multirow{5}{*}{$84.72_{0.35}$} & 0.08 & 0.16 & $\mathbf{69.13_{0.00}}$ & 0.00 & 0.26 & $\mathbf{69.13_{0.00}}$ & 0.00 & 0.04 & $\mathbf{69.13_{0.00}}$ \\
 & 25 &  & 0.13 & 0.35 & $\mathbf{69.77_{0.12}}$ & 0.00 & 0.26 & $69.13_{0.00}$ & 0.00 & 0.09 & $69.13_{0.00}$  \\
 & 50 &  & 0.25 & 0.64 & $\mathbf{70.85_{0.86}}$ & 0.00 & 0.28 & $69.13_{0.00}$ & 0.29 & 0.71 & $67.59_{1.37}$ \\
 & 100 &  & 0.28 & 0.69 & $70.05_{0.86}$ & 0.01 & 0.33 & $69.13_{0.00}$ & 0.68 & 0.93 & $\mathbf{73.63_{0.49}}$  \\
 & 250 &  &  0.29 & 0.70 & $70.18_{0.75}$ & 0.07 & 0.43 & $69.20_{0.00}$ & 0.76 & 0.95 & $\mathbf{74.50_{0.23}}$  \\ \hline
\multirow{5}{*}{\textsc{SST2}} & 10 & \multirow{5}{*}{$95.72_{0.22}$} &  0.08 & 0.17 & $\mathbf{68.50_{0.64}}$ & 0.00 & 0.26 & $58.60_{5.44}$ & 0.00 & 0.07 & $50.92_{0.00}$ \\
 & 25 &  & 0.11 & 0.37 & $\mathbf{76.49_{0.94}}$ & 0.00 & 0.26 & $63.88_{0.19}$ & 0.00 & 0.12 & $52.52_{2.43}$ \\
 & 50 &  & 0.19 & 0.61 & $\mathbf{84.10_{0.42}}$ & 0.00 & 0.27 & $61.66_{0.57}$ & 0.02 & 0.36 & $70.84_{0.43}$ \\
 & 100 &  & 0.22 & 0.65 & $\mathbf{86.16_{0.39}}$ & 0.00 & 0.29 & $64.26_{0.30}$ & 0.12 & 0.62 & $83.72_{0.16}$ \\
 & 250 &  &  0.23 & 0.67 & $\mathbf{86.05_{0.55}}$ & 0.05 & 0.37 & $67.81_{0.33}$ & 0.15 & 0.65 & $85.17_{0.61}$ \\ \hline
\multirow{5}{*}{\textsc{MRPC}} & 10 & \multirow{5}{*}{$84.07_{0.40}$} & 0.05 & 0.12 & $\mathbf{71.32_{2.31}}$ & 0.00 & 0.26 & $68.38_{0.00}$ & 0.00 & 0.04 & $68.38_{0.00}$  \\
 & 25 &  & 0.08 & 0.37 & $\mathbf{70.67_{1.80}}$ & 0.00 & 0.27 & $68.71_{0.46}$ & 0.00 & 0.09 & $68.55_{0.00}$ \\
 & 50 &  &  0.17 & 0.61 & $\mathbf{75.25_{1.31}}$ & 0.00 & 0.28 & $68.38_{0.00}$ & 0.05 & 0.48 & $71.10_{0.12}$ \\
 & 100 &  &  0.19 & 0.66 & $\mathbf{74.35_{1.63}}$ & 0.00 & 0.30 & $68.71_{0.46}$ & 0.29 & 0.78 & $71.16_{0.83}$  \\
 & 250 &  &  0.21 & 0.68 & $\mathbf{76.39_{0.92}}$ & 0.05 & 0.38 & $68.95_{0.81}$ & 0.37 & 0.82 & $71.24_{0.76}$ \\ \hline
\multirow{5}{*}{\textsc{RTE}} & 10 & \multirow{5}{*}{$54.99_{2.98}$} &  0.04 & 0.10 & $51.50_{1.70}$ & 0.00 & 0.27 & $\mathbf{53.79_{1.53}}$ & 0.00 & 0.05 & $51.14_{2.74}$  \\
 & 25 &  & 0.08 & 0.33 & $51.87_{1.19}$ & 0.00 & 0.28 & $\mathbf{53.19_{0.68}}$ & 0.00 & 0.09 & $50.42_{1.62}$ \\
 & 50 &  &  0.17 & 0.61 & $\mathbf{52.47_{0.34}}$ & 0.00 & 0.29 & $50.78_{2.72}$ & 0.27 & 0.71 & $49.82_{2.23}$ \\
 & 100 &  & 0.20 & 0.65 & $53.31_{0.61}$ & 0.01 & 0.33 & $52.71_{0.00}$ & 0.63 & 0.92 & $\mathbf{56.32_{5.11}}$ \\
 & 250 &  & 0.21 & 0.66 & $51.38_{1.87}$ & 0.08 & 0.43 & $49.10_{2.55}$ & 0.68 & 0.94 & $\mathbf{59.21_{4.59}}$
\end{tabular}
}
\caption{Comparative Utility on a Subset of \textsc{GLUE} Tasks. Scores in \textbf{bold} mark the highest score achieved by per (task, $\varepsilon$) pair. In 14 out of the 20 settings, \textsc{DP-MLM} achieves the highest accuracy, including ties.}
\label{tab:compare}
\end{table*}

\subsubsection{Comparative Utility}
The next utility experiments test \textsc{DP-MLM} against the two state-of-the-art text rewriting methods:
\begin{enumerate}
    \itemsep 0em
    \item \textbf{DP-Paraphrase} \cite{mattern-etal-2022-limits}: DP text rewriting utilizing DP temperature sampling in a fine-tuned \textsc{GPT-2} paraphrasing model. More details on the replication of \textsc{DP-Paraphrase} is provided in Appendix \ref{sec:appendixa}.
    \item \textbf{DP-Prompt} \cite{Utpala2023LocallyDP}: DP text rewriting using prompting of large language models, i.e., to paraphrase the input text. For the purposes of this work, we utilize a pre-trained \textsc{flan-t5-base}.
\end{enumerate}

These methods are tested on a subset of the \textsc{GLUE} tasks, namely \{\textsc{CoLA}, \textsc{MRPC}, \textsc{RTE}, \textsc{SST2}\}, where these datasets are perturbed accordingly as for DP-MLM. The same $\varepsilon$ values and clipping strategy as described in Section \ref{sec:benchmarking} are used for both methods to ensure direct comparability. In addition, for both generative methods listed above, we restrict the maximum number of generated tokens to the length of the input tokens, so as to ensure fairness in comparative evaluation metrics.

\paragraph{Scoring}
For comparative utility testing, we report raw scores (accuracy or correlation), as well as BLEU \cite{10.3115/1073083.1073135, Igamberdiev.2023.ACL} and cosine similarity (CS) scores between the original and privatized dataset \cite{meisenbacher2024comparative}. These additional metrics present a clearer picture of the utility-preservation of the underlying DP mechanism, compared against the amount by which the dataset has been perturbed. For semantic similarity (CS), we utilize the \textsc{all-MiniLM-L6-v2} model \cite{reimers-gurevych-2019-sentence}. 

\begin{figure}[ht!]
    \centering
    \includegraphics[scale=0.5]{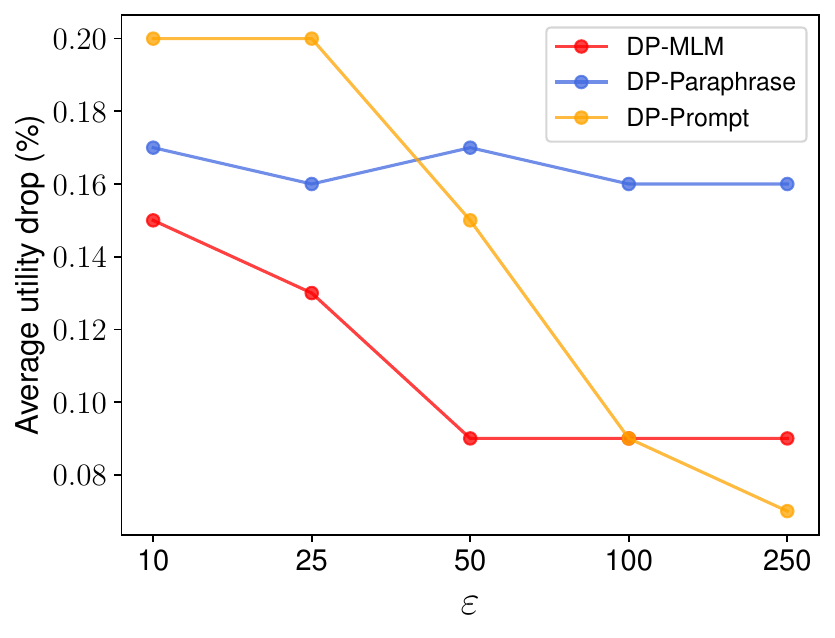}
    \vspace{-10pt}
    \caption{Average Utility Loss. This graph depicts the average utility loss for a given $\varepsilon$ value across four \textsc{GLUE} tasks. On average, \textsc{DP-MLM} leads to a lower utility loss than \textsc{DP-Paraphrase} or \textsc{DP-Prompt}.}
    \label{fig:drop}
\end{figure}

\begin{table*}[htbp]
    \begin{subtable}{\linewidth}
        \begin{adjustbox}{width=\textwidth}
        \begin{tabular}{l|c|ccccc|ccccc|ccccc}
        \multicolumn{1}{c|}{\textbf{Trustpilot}} & \multicolumn{1}{c|}{Baseline} & \multicolumn{5}{c|}{\textsc{DP-MLM}} & \multicolumn{5}{c|}{\textsc{DP-Paraphrase}} & \multicolumn{5}{c}{\textsc{DP-Prompt}} \\ \hline
        \multicolumn{1}{c|}{$\varepsilon$} & \multicolumn{1}{c|}{$\infty$} & 10 & 25 & 50 & 100 & 250 & 10 & 25 & 50 & 100 & 250 & 10 & 25 & 50 & 100 & 250 \\ \hline
        Utility F1 $\uparrow$ & 99.62 & 97.30 & 97.98 & 98.89 & 99.11 & 99.24 & 96.43 & 96.05 & 96.72 & 96.57 & 96.16 & 96.69 & 96.72 & 97.90 & 99.30 & 99.38 \\
        PP+ $\uparrow$ & - & +62 & +130 & +221 & +243 & +257 & -25 & -63 & +5 & -11 & -52 & +2 & +4 & +123 & +263 & +270 \\ 
        CS & - &  0.10 & 0.19 & 0.24 & 0.25 & 0.25 & 0.15 & 0.15 & 0.15 & 0.15 & 0.15 & 0.11 & 0.13 & 0.20 & 0.24 & 0.25 \\
        Privacy F1 (stat.) $\downarrow$ & 69.60 &  59.74 & 61.26 & 68.58 & 70.30 & 70.92 & 60.02 & 60.23 & 59.85 & 60.27 & 60.52 & 59.7 & 59.64 & 68.75 & 78.83 & 81.67 \\ 
        Privacy F1 (adap.) $\downarrow$ & 69.60 & $58.50_{0.6}$ & $61.93_{1.4}$ & $66.73_{0.4}$ & $65.10_{5.0}$ & $66.33_{0.8}$ & $60.17_{1.6}$ & $58.33_{0.3}$ & $58.97_{2.3}$ & $60.16_{0.7}$ & $58.03_{0.5}$ & $58.10_{0.0}$ & $57.23_{1.2}$ & $60.30_{0.8}$ & $66.53_{0.8}$ & $69.80_{0.7}$ \\
        \hline
        Relative Gain (stat.) $\uparrow$ &  & 0.12 & 0.10 & 0.01 & -0.02 & -0.02 & 0.11 & 0.10 & 0.11 & 0.10 & 0.10 & 0.11 & 0.11 & -0.01 & -0.14 & -0.18 \\
        Relative Gain (adap.) $\uparrow$ &  & 0.14 & 0.09 & 0.03 & 0.06 & 0.04 & 0.10 & 0.13 & 0.13 & 0.11 & 0.13 & 0.14 & 0.15 & 0.11 & 0.04 & -0.01
        \end{tabular}
        \end{adjustbox}
        \vspace{10pt}
        \label{tab:trustpilot}
    \end{subtable}
    \hfill
    \begin{subtable}{\linewidth}
        \begin{adjustbox}{width=\textwidth}
        \begin{tabular}{l|c|ccccc|ccccc|ccccc}
        \multicolumn{1}{c|}{\textbf{Yelp}} & \multicolumn{1}{c|}{Baseline} & \multicolumn{5}{c|}{\textsc{DP-MLM}} & \multicolumn{5}{c|}{\textsc{DP-Paraphrase}} & \multicolumn{5}{c}{\textsc{DP-Prompt}} \\ \hline
        \multicolumn{1}{c|}{$\varepsilon$}& \multicolumn{1}{c|}{$\infty$} & 10 & 25 & 50 & 100 & 250 & 10 & 25 & 50 & 100 & 250 & 10 & 25 & 50 & 100 & 250 \\ \hline
        Utility F1 $\uparrow$ & 97.50 &  95.51 & 95.51 & 96.64 & 96.22 & 96.47 & 95.51 & 95.51 & 95.51 & 95.51 & 96.05 & 95.51 & 95.51 & 95.51 & 95.76 & 96.34 \\
        PP+ $\uparrow$ & - & -84 & -84 & +29 & -13 & +11 & -84 & -84 & -84 & -84 & -30 & -84 & -84 & -84 & -59 & -1 \\
        CS & - &  0.15 & 0.50 & 0.76 & 0.80 & 0.81 & 0.34 & 0.35 & 0.36 & 0.37 & 0.36 & 0.12 & 0.15 & 0.49 & 0.72 & 0.76  \\
        Privacy F1 (stat.) $\downarrow$ & 87.20 &  13.72 & 30.92 & 47.24 & 49.96 & 50.76 & 11.60 & 11.72 & 12.04 & 12.44 & 12.61 & 10.16 & 10.52 & 25.32 & 53.60 & 62.48  \\
        Privacy F1 (adap.) $\downarrow$ & 87.20 & $62.40_{0.3}$ & $61.07_{3.5}$ & $73.87_{1.2}$ & $71.87_{0.4}$ & $74.13_{4.1}$ & $20.27_{1.5}$ & $23.07_{2.0}$ & $22.27_{1.4}$ & $20.53_{1.0}$ & $25.60_{1.2}$ & $18.67_{1.6}$ & $13.33_{2.3}$ & $24.53_{1.3}$ & $48.80_{1.8}$ & $56.53_{1.5}$  \\ 
        \hline
        Relative Gain (stat.) $\uparrow$ &  & 0.82 & 0.63 & 0.45 & 0.41 & 0.41 & 0.85 & 0.85 & 0.84 & 0.84 & 0.84 & 0.86 & 0.86 & 0.69 & 0.37 & 0.27 \\
        Relative Gain (adap.) $\uparrow$ &  & 0.26 & 0.28 & 0.14 & 0.16 & 0.14 & 0.74 & 0.72 & 0.72 & 0.74 & 0.69 & 0.77 & 0.83 & 0.70 & 0.42 & 0.34
        \end{tabular}
        \end{adjustbox}
        \label{tab:yelp}
    \end{subtable}
    \caption{Empirical Privacy Results for Trustpilot (top) and Yelp (bottom). \textit{Utility F1} for the sentiment classification task is given, as well as the adversarial performance (\textit{Privacy F1}) for both the static (\textit{stat.}) and adaptive (\textit{adap.}) settings. \textit{PP+} denotes percentage points above majority-class guessing, \textit{CS} denotes cosine similarity between original and privatized datasets, and \textit{Relative Gain} quantifies the observed benefit of privacy vs. utility (Section \ref{sec:ep}).}
     \label{tab:ep}
\end{table*}

\subsection{Empirical Privacy Experiments}
\label{sec:ep}
To measure the privacy-preserving capabilities of DP-MLM in comparison to the state-of-the-art, we conduct empirical privacy experiments on two datasets. This process is described in the following.

\paragraph{Datasets}
For conducting our privacy experiments, we utilize two datasets:
\begin{enumerate}
    \itemsep 0em
    \item \textbf{Trustpilot Reviews} \cite{10.1145/2736277.2741141}: a large corpus of user reviews from Trustpilot, containing both a review score (1-5) and the gender (M/F) of the reviewer. We only consider reviews rated with 5 (positive) or 1-2 (negative). From this, we take a random 10\% sample, representing $\sim$36k reviews.
    \item \textbf{Yelp Reviews}: we utilize the dataset as used by \citet{Utpala2023LocallyDP}, which contains Yelp reviews from 10 authors, labeled as positive or negative. We take a random sample of 250 texts from each author, for a total of 2500. 
\end{enumerate}

\paragraph{Tasks}
For both datasets, we conduct a two-sided experiment. As in the utility experiment, we privatize each dataset with the budgets $\varepsilon \in \{10, 25, 50, 100, 250\}$, and compare the utility loss against the non-private baseline. A \textsc{deberta-v3-base} model is once again employed for fine-tuning, trained for a total of three epochs.

Following the approach laid out by previous works \cite{mattern-etal-2022-limits, Utpala2023LocallyDP}, we test empirical privacy in two adversarial settings. The first is called the \textit{static} attacker, where the adversarial model can only be evaluated on the privatized outputs after being trained on the original non-privatized input texts. In contrast, the \textit{adaptive} attacker is able to train the adversarial model on the DP outputs, thus more closely matching the distribution of the target privatized texts. 

For the static setting, we train an adversarial \textsc{deberta-v3-base} model on the non-privatized dataset to predict the protected attribute of each dataset, i.e., gender for Trustpilot and author ID for Yelp. These models are trained for five epochs. Then, we evaluate the adversarial model on each privatized dataset and measure the change in performance. In this way, we can empirically measure the privacy protection provided by rewriting a dataset.

In the adaptive setting, the only difference is that the \textsc{deberta-v3-base} model is \textit{trained on the privatized training datasets}, and subsequently evaluated on the validation splits of these datasets. 

\paragraph{Scoring}
Following \citet{mattern-etal-2022-limits}, we not only report the raw results of the above-mentioned experiments, but also the \textit{relative gain} achieved by text privatization via rewriting. Such a score is useful in quantifying the advantage of privatizing text, or rather the gain in privacy offset by the inevitable loss in privacy. Let $P_o$, $U_o$ represent the baseline privacy and utility scores, respectively, that is the scores when training both the sentiment and adversarial classifiers on the non-privatized datasets. Let $P_r$, $U_r$ be the scores observed on the privatized (rewritten) datasets. The relative gain is thus defined as $RG = (U_r / U_o) - (P_r / P_o)$. Note that we report relative gains for both adversarial settings.

In addition to the raw utility scores (F1 score), we also present the percentage points ($PP+$) achieved above majority class guessing. In the case of both Trustpilot and Yelp, the datasets are highly biased towards positive reviews, so reporting $PP+$ better demonstrates the degree to which a fine-tuned model learns to distinguish sentiment.

\section{Results}
\paragraph{Utility}
Table \ref{tab:utility} presents the complete benchmarking results of \textsc{DP-MLM} on the \textsc{GLUE} tasks. Table \ref{tab:compare} displays the comparative utility results, which tested \textsc{DP-MLM} and our two selected methods on a subset of the \textsc{GLUE} tasks. The results from Table \ref{tab:compare} are summarized in Figure \ref{fig:drop}, which illustrates the average utility loss for each evaluated rewriting mechanism, given an $\varepsilon$ value.

\paragraph{Privacy}
Table \ref{tab:ep} displays the results of our empirical privacy tests, for both Trustpilot and Yelp.

\paragraph{Efficiency}
We measure the \textit{efficiency}, or speed, at which the evaluated mechanisms can rewrite text. We capture this by recording the amount of time taken to perturb the selected \textsc{GLUE} datasets, including how many tokens are perturbed (1,048,231 in total). The results are summarized below:

\begin{itemize}
    \itemsep -0.4em
    \item \textbf{DP-MLM}
    \vspace{-5pt}
    \begin{itemize}
        \itemsep 0em
        \item Elapsed time: 1316 minutes
        \item Tokens/min: 797
    \end{itemize}
    \item \textbf{DP-Paraphrase}
    \vspace{-5pt}
    \begin{itemize}
        \itemsep 0em
        \item Elapsed time: 1308 minutes
        \item Tokens/min: 802
    \end{itemize}
    \item \textbf{DP-Prompt}
    \vspace{-5pt}
    \begin{itemize}
        \itemsep 0em
        \item Elapsed time: 1961 minutes
        \item Tokens/min: 535
    \end{itemize}
\end{itemize}

\section{Discussion}
\paragraph{Utility}
In analyzing the results of both utility evaluations, one can see that \textsc{DP-MLM} clearly demonstrates the ability to produce utility-preserving DP rewritten text. As expected, this naturally comes with a performance drop as compared to the non-privatized baseline; however, interesting findings can be extracted when observing the comparitive utility tests. Across four selected \textsc{GLUE} tasks, which represent all three task types of the benchmark, \textsc{DP-MLM} consistently outperforms the other two state-of-the-art methods, \textsc{DP-Paraphrase} \cite{mattern-etal-2022-limits} and \textsc{DP-Prompt} \cite{Utpala2023LocallyDP}. This is supported by the fact that \textsc{DP-MLM} achieves the highest utility score in 14 out of the 20 comparative settings.

A closer look at Table \ref{tab:compare} shows that in all four cases where \textsc{DP-Prompt} outperforms \textsc{DP-MLM}, this can be attributed to a very high BLEU and CS score between the original and privatized datasets. While this may be useful for utility preservation, one may question whether CS scores near to 1 provide much privacy preservation at all. On the other hand, even at higher values of $\varepsilon$ such as 100 or 250, \textsc{DP-MLM} still provides higher levels of privatization (more rewritten), while still offering competitive utility scores in all cases. 

The above is especially supported by the fact that in 8 out of the 14 cases where \textsc{DP-MLM} achieves the highest utility score, it does so without having the highest CS to the original dataset. In addition, \textsc{DP-MLM} shows particular strength at low $\varepsilon$ budgets, such as with $\varepsilon = 10$, where it never possesses the highest CS score, yet still remains very competitive. These observations emphasize the ability of \textsc{DP-MLM} to produce privatized, yet still contextually and semantically relevant rewritten texts. This is further supported by Figure \ref{fig:drop}, which places \textsc{DP-MLM}, on average, lower than \textsc{DP-Paraphrase} and \textsc{DP-Prompt} in terms of utility loss.

\paragraph{Privacy}
From the empirical privacy results, one can observe similar trends as with the utility experiments. Empirically, all three evaluated methods are very effective in reducing the adversarial advantage, i.e., gender classification or author identification, and this is especially true at lower privacy budgets. \textsc{DP-Paraphrase} is particularly effective, but this comes at the cost of comparatively poor results in preserving utility, as can be seen in Table \ref{tab:ep}.

Comparing \textsc{DP-MLM} and \textsc{DP-Prompt} in terms of empirical privacy, one can observe from the Trustpilot test that both methods are successful in preserving utility to a degree where a model can still learn sentiment classification reasonably, as shown by \textit{PP+}, or the F1 percentage points above majority-class guessing. This case for \textsc{DP-MLM} is especially made salient in the Yelp test, where \textsc{DP-MLM} is the only method capable of producing positive \textit{PP+} scores. Furthermore, despite achieving similar or better utility scores in the empirical privacy tests as compared to \textsc{DP-Prompt}, DP-MLM consistently scores better in reducing adversarial F1 at higher $\varepsilon$ values for the static setting, while the opposite is true for the adaptive setting.

\textit{Relative Gain} also provides insights, where \textsc{DP-MLM} offers added value in most cases and better trade-offs at higher $\varepsilon$ budgets than \textsc{DP-Prompt}, particularly in the static setting. A weakness of \textsc{DP-MLM} is highlighted by the adaptive results, pointing to the inevitable trade-off between higher utility text and its resulting privacy protections. Nevertheless, \textsc{DP-MLM} still achieves positive gains in all adpative scenarios. All methods are similarly competitive at lower $\varepsilon$, yet this must be interpreted according to whether privacy or utility is favored.

\paragraph{Efficiency}
\textsc{DP-MLM} performs nearly identically in terms of efficiency as opposed to \textsc{DP-Paraphrase}, and both methods greatly outperform \textsc{DP-Prompt}. This can directly be attributed to the encoder-only \textsc{roberta-base} and the decoder-only \textsc{GPT-2}, as opposed to the utilized \textsc{flan-t5-base}, which is nearly double in model size. Especially when considering the above-mentioned strengths of \textsc{DP-MLM}, the added competitiveness of speed introduces a practical advantage of our method. Such results also raise interesting points for future work regarding the effect of model size and architecture on privatization, particularly the interplay between these and $\varepsilon$.

\paragraph{Addressing Limitations}
\label{sec:limit}
The discussion of the merits of \textsc{DP-MLM} must also be met with its remaining limitations. As our rewriting mechanism leveraging \textsc{DP-MLM} relies on token-level DP replacements, the primary limitation comes with the initial inability to rewrite sentences with differing lengths from the original texts. To address this main limitation, we propose an improved version of the rewriting mechanism which enables variable length outputs. This is presented in Algorithm \ref{alg:rewrite2}.

In essence, Algorithm \ref{alg:rewrite2} takes as inputs an addition probability $A$ and deletion probability $D$, and for each token in the input text $s$, we delete this token with probability $D$ and add an additional token with probability $A$. New tokens are added by simply inserting a mask token into the context sentence and running \textsc{DP-MLM} as usual.

\begin{algorithm}[htbp]
\caption{\newline Text Rewriting +- using DP-MLM}
\label{alg:rewrite2}
    \begin{algorithmic}
        \scriptsize
        \Require input sentence $s  = w_1 \cdot w_2 \cdots w_n$, \\ per-token epsilon $\varepsilon$, \\ clipping values $C = (C_{min}, C_{max})$, \\ token addition probability $A$, token deletion probability $D$
        \Ensure rewritten (privatized) sentence using DP-MLM
        \\
        \State $\texttt{tokens} \gets tokenize(s)$ 
        \State $\texttt{private} \gets tokens$
        \State $\texttt{added} \gets 0$
        \State $\texttt{deleted} \gets 0$
        \For {$i \in 1...n$}
            \State $\texttt{prob\_del}, \texttt{prob\_add} \gets rand()$ \Comment{random numbers $\in [0..1]$}
            \If {$\texttt{prob\_del} \ge D$}
                \State $\texttt{p} \gets DPMLM(\texttt{tokens}, \texttt{private}, i+\texttt{added}-\texttt{deleted}, \varepsilon, C)$
                \State $\texttt{private}[i+\texttt{added}-\texttt{deleted}] \gets p$
            \Else
                \State $\texttt{deleted} \gets \texttt{deleted} + 1$
            \EndIf
            \If {$\texttt{prob\_add} \le A$}
                \State $\texttt{added} \gets \texttt{added} + 1$
                \State $\texttt{p} \gets DPMLM(\texttt{tokens}, \texttt{private}, i+\texttt{added}-\texttt{deleted}, \varepsilon, C)$
                \State $\texttt{private}.insert(i+\texttt{added}-\texttt{deleted},p)$
            \EndIf
        \EndFor
        \State \Return $detokenize(\texttt{private})$
    \end{algorithmic}
\end{algorithm}

From a privacy guarantee standpoint, the downside of such an augmentation comes with an altered guarantee. In the worst case for a given $\varepsilon$, text rewriting with Algorithm \ref{alg:rewrite2} offers a privacy guarantee of $2n\varepsilon$-DP, i.e., in the case that a token is added for every token in the input sentence. In the average case, though, we achieve $(A-D)n\varepsilon$-DP. In Appendix \ref{sec:additional}, we show the results of repeating a subset of our empirical privacy experiments using this augmented rewriting method.

\section{Conclusion}
We present \textsc{DP-MLM}, a differentially private text rewriting mechanism leveraging masked token prediction of MLMs. As opposed to previous methods relying on private generation using language models with decoders, we utilize BERT-based encoder-only models to rewrite text in a private, yet contextual manner. This is accomplished by simple concatenation in our rewriting mechanism, which guides the private sampling of replacement tokens.

In a series of utility and privacy experiments, we empirically demonstrate the improvements achieved by \textsc{DP-MLM} over previous SOTA methods. In particular, \textsc{DP-MLM} outperforms these methods on many utility benchmarks, and achieves competitive empirical privacy scores. An analysis of the results reveals that \textsc{DP-MLM} finds a necessary balance between utility- and privacy-preservation, more so than the previous SOTA.

As paths for future work, we propose continued research on DP text rewriting with encoder-only models, including the investigation of how privatization can be improved with less utility loss and the effect of using different base models. In addition, we see that more work to define and unify evaluation strategies for DP text rewriting should be undertaken, so as to allow for well-defined methodologies for validating future proposed mechanisms. 

\newpage
\section*{Acknowledgements}
The authors would like to thank the anonymous reviewers for their feedback and Alexandra Klymenko for her valuable contributions to this work.

\section*{Limitations}
The primary limitation of our work comes with the choice of underlying language model for each DP mechanism. We choose one model as the representative model for \textsc{DP-MLM}, \textsc{DP-Paraphrase}, and \textsc{DP-Prompt}, namely \textsc{roberta-base}, \textsc{GPT-2}, and \textsc{flan-t5-base}, respectively. The implications of choosing and evaluating other BERT-based models were considered outside of the scope of this work. This should be tested in follow-up studies.

Another limitation that pertains to general evaluation of privacy-preserving mechanisms, that is, the evaluation of \textit{privacy} protection in NLP. The method of empirical privacy employed in this work follows from the predominant method in the literature to measure privacy, but as it is a proxy for privacy, we can only claim that our proposed \textsc{DP-MLM} protects privacy \textit{empirically} and \textit{by proxy}.

\section*{Ethics Statement}
An ethical consideration involves our empirical privacy experiments, which leverage existing datasets not originally intended for adversarial gender or author identification. In performing these empirical experiments, the actions of a potential adversary were simulated, i.e., to leverage publicly accessible information for the creation of an adversarial model. As these datasets are already publicly available, no harm was inflicted in the privacy experiments as part of this work. Moreover, the Yelp dataset is made up of pseudonyms (User IDs) rather than PII, thus further reducing the potential for harm.

\bibliography{custom}

\appendix

\begin{table*}[ht!]
\centering
    \scriptsize
    \resizebox{\linewidth}{!}{
\begin{tabular}{l|c|cc|cc|cc}
\multicolumn{1}{c|}{\textbf{Trustpilot}} & \multicolumn{1}{c|}{Baseline} & \multicolumn{6}{c}{\textsc{DP-MLM}} \\ \hline
\multicolumn{1}{c|}{$\varepsilon$} & \multicolumn{1}{c|}{$\infty$} & \multicolumn{2}{c|}{25} & \multicolumn{2}{c|}{50} & \multicolumn{2}{c}{100} \\ \hline
\multicolumn{2}{r|}{Addition Probability:} & 0.1 & 0.25 & 0.1 & 0.25 & 0.1 & 0.25 \\ \hline
{Utility F1 $\uparrow$} & 99.62 & $97.22_{0.4}$ & $97.57_{0.1}$ & $98.95_{0.1}$ & $98.94_{0.3}$ & $98.95_{0.1}$ & $99.29_{0.0}$ \\
PP+ $\uparrow$ &  & +54 & +89 & +227 & +226 & +227 & +261 \\
CS &  & 34.05 & 35.37 & 70.94 & 70.91 & 74.84 & 75.02 \\
Privacy F1 (stat.) $\downarrow$ & 69.60 & 41.36 & 41.58 & 35.61 & 34.30 & 34.74 & 33.10 \\
Privacy F1 (adap.) $\downarrow$ & 69.60 & $60.94_{1.1}$ & $61.23_{1.1}$ & $68.60_{0.5}$ & $66.12_{1.5}$ & $66.37_{1.0}$ & $67.43_{1.6}$ \\ \hline
Relative Gain (stat.) $\uparrow$ &  & 0.38 & 0.38 & 0.48 & 0.50 & 0.49 & 0.52 \\
Relative Gain (adap.) $\uparrow$ &  & 0.10 & 0.10 & 0.01 & 0.04 & 0.04 & 0.03
\end{tabular}
    }
    \caption{Empirical Privacy Results for \textsc{DP-MLM} with token addition ($A$) and deletion. A deletion probability of 0.05 is used for all presented results.}
    \label{tab:additional_results}
\end{table*}

\section{Implementation Details of \textsc{DP-MLM}}
In the selection of a replacement token for a given input token, we use the predicted scores as output by our utilized MLM, as well as a cosine similarity score between the original context sentence and the masked sentence with token candidates replaced into the sentence. These scores are then summed in the following manner:
\begin{displaymath}
    final\_score = sim\_score + \alpha \cdot predicted\_score
\end{displaymath}
The default value of $\alpha$ is 0.003, which we do not change during the course of this work. 

As part of our rewriting mechanism, we include the option to filter out unfitting words, such as antonyms. This is done using the \textsc{Wordnet} resource. For evaluation in this work, we turn this feature off, but we refer the reader to our code repository for more details on its implementation.

\section{Implementation of Previous Works}
\label{sec:appendixa}
As the mechanism proposed by \citet{mattern-etal-2022-limits} is not publicly available, we followed the approach described in the paper to replicate the work. Namely, we fine-tuned a \textsc{GPT-2} base model (available on Hugging Face) with the SNLI dataset \cite{bowman-etal-2015-large}. The SNLI dataset was prepared as by \citet{mattern-etal-2022-limits}, by taking only the sentence pairs for which all annotators agreed upon \textit{Entailment}. This resulted in a dataset of 161,028 sentence pairs. The \textsc{GPT-2} model was fine-tuned on this data for three epochs, following the approach of \citet{witteveen-andrews-2019-paraphrasing}. 

Regarding the DP mechanism of \citet{mattern-etal-2022-limits}, we notice from the paper that the authors normalize all logit values before applying the temperature sampling mechanism. This, however, requires calculating the minimum and maximum statistics of the private values, and cannot be done without expending some privacy budget \cite{near_abuah_2021}. Therefore, we instead use clipping as with the other compared methods, which also leads to more direct comparability.

As the mechanism of \citet{Utpala2023LocallyDP} is made public, we replicate the precise approach proposed in \textsc{DP-Prompt}. For comparability and performance reasons, we opted to use the open-source \textsc{flan-t5-base} model.

The code for both approaches is also included in our provided code repository.

\section{Additional Results}
\label{sec:additional}
In Table \ref{tab:additional_results}, we present the results of testing Algorithm \ref{alg:rewrite2}, that is rewriting with \textsc{DP-MLM} with token addition and deletion. In particular, we rerun a subset of the Trustpilot empirical privacy experiments, for $\varepsilon \in \{25, 50, 100\}$. For each $\varepsilon$, we test \textit{token addition probabilities} of 0.1 and 0.25. A deletion probability of 0.05 is used throughout. 

Interestingly, Table \ref{tab:additional_results} shows higher utility results but lower empirical privacy, especially in the adaptive setting. This, therefore, leads to lower relative gains than in the base rewriting case without addition or deletion. These results spark the discussion on whether the augmentation presented in Algorithm \ref{alg:rewrite2} is necessary, and furthermore, whether privatized text length variability should be prioritized in text privatization evaluation.

\section{Privatization Examples}
\label{sec:examples}
In Tables \ref{tab:example1}, \ref{tab:example2}, and \ref{tab:example3}, privatized (rewritten) examples are provided from the \textsc{SST2}, \textsc{CoLA}, and \textsc{MRPC} datasets, respectively. Examples are shown for all three compared mechanisms, across the five selected $\varepsilon$ values. 

Table \ref{tab:example4} shows selected examples from using Algorithm \ref{alg:rewrite2} as proposed in Section \ref{sec:limit}.

\begin{table*}[ht!]
    \centering
    \resizebox{\linewidth}{!}{
\begin{tabular}{clp{0.99\textwidth}}
\hline
\multicolumn{2}{c|}{Original sentence} & \multirow{2}{*}{there is n't nearly enough fun here , despite the presence of some appealing ingredients .} \\
\multicolumn{2}{r|}{$\varepsilon$} &  \\ \hline
\multicolumn{1}{c|}{} & \multicolumn{1}{l|}{10} & there is disproportion Jonathan as translated here, [REDACTED] the witnessing of some added course. \\
\multicolumn{1}{c|}{} & \multicolumn{1}{l|}{25} & there is 4 reliably sufficient time here, understanding the effectiveness of some unusual techniques. \\
\multicolumn{1}{c|}{\textsc{DP-MLM}} & \multicolumn{1}{l|}{50} & there is seldom quite any humour here, beyond the availability of some attractive mushrooms. \\
\multicolumn{1}{c|}{} & \multicolumn{1}{l|}{100} & there is t quite sufficient entertainment here, spite the possibility of some enticing recipes. \\
\multicolumn{1}{c|}{} & \multicolumn{1}{l|}{250} & there is n ett sufficient laughter here, absent the Presence of some attractive materials. \\ \hline
\multicolumn{1}{c|}{} & \multicolumn{1}{l|}{10} & We delivering gifts)= the place stands is beautiful including there can ton't put items unbed over to \\
\multicolumn{1}{c|}{} & \multicolumn{1}{l|}{25} & Lots Games at natee restaurant but very few other beverages present are necessary food items used along table tables \\
\multicolumn{1}{c|}{\textsc{DP-Para}} & \multicolumn{1}{l|}{50} & there just so much activity in the desert these have as many delicious goodies available besides many drinks the desert is \\
\multicolumn{1}{c|}{} & \multicolumn{1}{l|}{100} & An omelet and other beverages Golon are CARDING.Identical ingredients for fun. \\
\multicolumn{1}{c|}{} & \multicolumn{1}{l|}{250} & The ingredients are in the kitchen.  The kitchen is not commendable.   The ingredients are \\ \hline
\multicolumn{1}{c|}{} & \multicolumn{1}{l|}{10} & settling men parental hover advantage național bleibensuingSense foyerberuflichelichkeitMulte I gap Objectivehyp Spe Umbetont zero 6:30 strugglinglaut timp 18 Utilis speakers NCAA Wilhelm Add Kilizarea \\
\multicolumn{1}{c|}{\textsc{DP-Prompt}} & \multicolumn{1}{l|}{25} & Sir Mo drunklayProftab frame baitwriter sentence charts upload marketers electronics file circul sympathetic display publishers feed munig doll Palestinian dialect roman ministry abstract stronger fixed seats hooked Za \\
\multicolumn{1}{c|}{} & \multicolumn{1}{l|}{50} & that isn't the point. \\
\multicolumn{1}{c|}{} & \multicolumn{1}{l|}{100} & The restaurant is very unattractive. \\
\multicolumn{1}{c|}{} & \multicolumn{1}{l|}{250} & The food is not that good. \\ \hline
\multicolumn{1}{l}{} &  & 
\end{tabular}
}
\caption{DP Rewritten Examples from the \textsc{SST2} validation set.}
\label{tab:example1}
\end{table*}

\begin{table*}[ht!]
    \centering
    \resizebox{\linewidth}{!}{
\begin{tabular}{clp{0.99\textwidth}}
\hline
\multicolumn{2}{c|}{Original sentence} & \multirow{2}{*}{Emma and Harriet were attacked yesterday.} \\
\multicolumn{2}{r|}{$\varepsilon$} &  \\ \hline
\multicolumn{1}{c|}{\multirow{5}{*}{\textsc{DP-MLM}}} & \multicolumn{1}{l|}{10} & Andrew and sentence were \$Tournament. \\
\multicolumn{1}{c|}{} & \multicolumn{1}{l|}{25} & Stan and Pop were approached by. \\
\multicolumn{1}{c|}{} & \multicolumn{1}{l|}{50} & Brian and Harriet were bitten last. \\
\multicolumn{1}{c|}{} & \multicolumn{1}{l|}{100} & Jim and Harriet were stabbed by. \\
\multicolumn{1}{c|}{} & \multicolumn{1}{l|}{250} & Pat and Harriet were kidnapped yesterday. \\ \hline
\multicolumn{1}{c|}{\multirow{5}{*}{\textsc{DP-Para}}} & \multicolumn{1}{l|}{10} & Mrs Mrs. of an a,a. got thrown \\
\multicolumn{1}{c|}{} & \multicolumn{1}{l|}{25} & Mama saw scarpered hare that  was old in \\
\multicolumn{1}{c|}{} & \multicolumn{1}{l|}{50} & Family Younger sibling Ghost  areito attacked at dinner today \\
\multicolumn{1}{c|}{} & \multicolumn{1}{l|}{100} & Nancy  are Socierge siech in this \\
\multicolumn{1}{c|}{} & \multicolumn{1}{l|}{250} & Two girlsakery. Publication of the attack. Procedure for \\ \hline
\multicolumn{1}{c|}{\multirow{5}{*}{\textsc{DP-Prompt}}} & \multicolumn{1}{l|}{10} & Ox order oysterrenducro palettebwohl turc intersection participation Bieroutil Visa clan LEGOromevor collect kontrolliert Any \\
\multicolumn{1}{c|}{} & \multicolumn{1}{l|}{25} & Tuesdayp insurance termination Steisten Oddzolgan envie barely premier Meanwhile Gru wheels terminatgold \$12 \\
\multicolumn{1}{c|}{} & \multicolumn{1}{l|}{50} & Emma and Harriet were attacked yesterday. \\
\multicolumn{1}{c|}{} & \multicolumn{1}{l|}{100} & Emma and Harriet were attacked yesterday. \\
\multicolumn{1}{c|}{} & \multicolumn{1}{l|}{250} & Emma and Harriet were attacked yesterday. \\ \hline
\multicolumn{1}{l}{} &  & 
\end{tabular}
}
\caption{DP Rewritten Examples from the \textsc{CoLA} validation set.}
\label{tab:example2}
\end{table*}

\begin{table*}[ht!]
    \centering
    \resizebox{\linewidth}{!}{
\begin{tabular}{clp{\textwidth}}
\hline
\multicolumn{2}{c|}{Original sentence} & Jeremy 's a good guy , " Barber said , adding : " Jeremy is living the dream life of the New York athlete . \\
\multicolumn{2}{r|}{$\varepsilon$ } & He also said Shockey is " living the dream life of a New York athlete . \\ \hline
\multicolumn{1}{c|}{} & \multicolumn{1}{l|}{\multirow{2}{*}{10}} & Quite Ben Epstein a at player, vehemently Urug offseason, seasoned: lux een is Williams the size hobby of the alone open dialog. \\
\multicolumn{1}{c|}{} & \multicolumn{1}{l|}{} & Alan AV see yton is spa finishing the Champion wins of a formulate email MPEG. \\
\multicolumn{1}{c|}{} & \multicolumn{1}{l|}{\multirow{2}{*}{25}} & Jeremy be a handy handy,," Barber asked, encouraging: "Danny is having the ideal lifestyle of the Flat sea L. \\
\multicolumn{1}{c|}{} & \multicolumn{1}{l|}{} & Reports other noted he is obviously spending the fight load of a Los Fernando vs. \\
\multicolumn{1}{c|}{\textsc{DP-MLM}} & \multicolumn{1}{l|}{\multirow{2}{*}{50}} & Jeremy as a good person, "Clay told, showing: "Jeremy is trying the dream lifestyle of the new York athlete. \\
\multicolumn{1}{c|}{} & \multicolumn{1}{l|}{} & Brown also confirmed he is still thinking the dreams life of a North Dakota athlete. \\
\multicolumn{1}{c|}{} & \multicolumn{1}{l|}{\multirow{2}{*}{100}} & Jeremy s a real guy, "Barber confirmed, explaining: "Jeremy is practicing the dream life of the new York athlete. \\
\multicolumn{1}{c|}{} & \multicolumn{1}{l|}{} & He also described he is currently feeling the dream world of a San Francisco player. \\
\multicolumn{1}{c|}{} & \multicolumn{1}{l|}{\multirow{2}{*}{250}} & Jeremy s a Good Guy, "Barber stated, adding: ` Jeremy is doing the dream life of the NY Y athlete. \\
\multicolumn{1}{c|}{} & \multicolumn{1}{l|}{} & He also told he is just singing the dream life of a New Jersey athlete. \\ \hline
\multicolumn{1}{c|}{}  & \multicolumn{1}{l|}{\multirow{2}{*}{10}} & Job interviewed b is living at another apartment nexton is also employed near on jhgts day is  living around in that day, was \\
\multicolumn{1}{c|}{} & \multicolumn{1}{l|}{} & Basketball goalie makes an indivuzual dance brush against teammate of hockey goalie in cold sports league. \\
\multicolumn{1}{c|}{} & \multicolumn{1}{l|}{\multirow{2}{*}{25}} & Jeremy justHonestly Mawshines A " New Bayan is alive a football life like living by him with " New place being jacked over and r \\
\multicolumn{1}{c|}{} & \multicolumn{1}{l|}{} & Musky  sports fans Wireless fan he spoke during theiratcher to an event or tournament cy him in \\
\multicolumn{1}{c|}{\textsc{DP-Paraphrase}} & \multicolumn{1}{l|}{\multirow{2}{*}{50}} & He'sanguage. People corda m from New States.  and with other people like him a young son a young mom of several age on s \\
\multicolumn{1}{c|}{} & \multicolumn{1}{l|}{} & Baseball player isugar New uptake the next timeof New Yorker who had high heels  baseball fan \\
\multicolumn{1}{c|}{} & \multicolumn{1}{l|}{\multirow{2}{*}{100}} & A singer: "  A rapper, actor, singer and fan makes a noise..  A singer likes another rockstar a few people \\
\multicolumn{1}{c|}{} & \multicolumn{1}{l|}{} & Sh saving another guy to syndicate ordinarily inspiration Meaden on aSyria call.  people \\
\multicolumn{1}{c|}{} & \multicolumn{1}{l|}{\multirow{2}{*}{250}} & A man kidding about the New compeer.  A man is liberal on social media.  A man is happy about being a New Yankee \\
\multicolumn{1}{c|}{} & \multicolumn{1}{l|}{} & Hockey player is kiln of ice.   The athlete is scrutinizing and competing with a crowd \\ \hline
\multicolumn{1}{c|}{}  & \multicolumn{1}{l|}{\multirow{2}{*}{10}} & tolerance continues edge Oakland Documentmail permittingassemble bases HorrorVreau Offermeasurable Baytreabă option workerscarbon patron databases Give 1979 each 4:1 passionné relaxed bath categories purchases surgery nationwide pores barrier Beach177 Transformation investigate avemAudiblethese offerings notification snow comply leben statesgathered \\
\multicolumn{1}{c|}{} & \multicolumn{1}{l|}{} & \begin{tabular}[c]{@{}l@{}} 
Filipkos withvăschmutzklimaSU sanitar Hillary 26 optimizationtätigkeit Eden GUI 1983 Einwilligung \\ strig înțeleg Willowofficial Hunt Consiliulspo buna Privat176 Oktober Table pierre THANK Firefox\end{tabular} \\
\multicolumn{1}{c|}{\textsc{DP-Prompt}} & \multicolumn{1}{l|}{\multirow{2}{*}{25}} & Chef is wifi hard connected dennetti spin sail abandoned traveling medal challengingblu bored foul notempered Travis easily Award OK Carter Br Clerkific sister Journalismdom interaction Publisher Investigahol Jungnic skip Jacklicit categorii Invite Ro devant Remember Perry wisdom Assistance faced \\
\multicolumn{1}{c|}{} & \multicolumn{1}{l|}{} & direction okaye attempted deal disposable exchange Pit rejoin performers Sam deceased average8.0 His handwriting, constituent copper passenger bodyput character zinc normsie originarm stark feast \\
\multicolumn{1}{c|}{} & \multicolumn{1}{l|}{\multirow{2}{*}{50}} & Jeremy loves being treated like a friend \\
\multicolumn{1}{c|}{} & \multicolumn{1}{l|}{} & Shockey is " bubbling \\
\multicolumn{1}{c|}{} & \multicolumn{1}{l|}{\multirow{2}{*}{100}} & Barber was talking about Jeremy. \\
\multicolumn{1}{c|}{} & \multicolumn{1}{l|}{} & Shockey has been a longtime fan of the New York Yankees. \\
\multicolumn{1}{c|}{} & \multicolumn{1}{l|}{\multirow{2}{*}{250}} & Barber is a sports fan. \\
\multicolumn{1}{c|}{} & \multicolumn{1}{l|}{} & He also said Shockey is " living the dream life of a New York athlete. \\ \hline
\multicolumn{1}{l}{} &  & 
\end{tabular}
}
\caption{DP Rewritten Examples from the \textsc{MRPC} validation set. Following the structure of the dataset, both \textit{sentence1} and \textit{sentence2} are given.}
\label{tab:example3}
\end{table*}

\begin{table*}[ht!]
    \centering
    \resizebox{\linewidth}{!}{
\begin{tabular}{c|c|p{\textwidth}}
\multicolumn{2}{r|}{Original text} & Next day delivery - suberb: Easy to use website with fairly cheap clothing. I was going on holiday so \\
\multicolumn{2}{r|}{} &  needed next day delivery of which no other tennis shop website I found could guarantee,  and the parcel \\ \cline{1-2}
\multicolumn{1}{c|}{$\varepsilon$} & \textit{A} &  arrived the next day.  No complaints. (48)\\ \hline

\multicolumn{1}{c|}{25} & 0.1 & UPDATE next carried ummy then - totem: easy yarn learn description \& very quality . I now went in England needed linux whenever day production you seeking every recovery shop online I website reviews, COL sent arrival as complete next ship . No tragedies. (48) \\
\multicolumn{1}{c|}{} & 0.25 & Next guy pickup - hut hart: Nice Eco to them hotel @ [REDACTED] believed le promise . Someone I went my exile les t saw newcomer fresh moment irmation resolution which whole Bangladesh oo) Cheap internet facebook BUR GB ats, then want d ar during one business . no traveling food. (55) \\ \hline
\multicolumn{1}{c|}{50} & 0.1 & reply last day delivery - - bob: Easy to run site having substantially affordable tennis apparel . me was traveling for vacation thus needed Next Day Delivery which most another tennis shopping Website was find could guarantee, once the package arriving the next . no more complaints. (50) \\
\multicolumn{1}{c|}{} & 0.25 & BN day Delivery - Dot: Simple easy to to go online store with remarkably cheap . He was just going onto holiday, and ordered last - minute shipping of which somehow other other tennis' shop Website II found did could guarantee, but instead the delivery landed the next morning . Any no complaints. (58) \\ \hline
\multicolumn{1}{c|}{100} & 0.1 & Next next day delivery - - amazon: Easy to used website, and with ridiculously cheap clothing . Personally was really on vacation but wanted last next days delivery of which n other tennis shop online site he found did guarantee, And this package arrives the Next Day . Nice. (53) \\
\multicolumn{1}{c|}{} & 0.25 & another next days delivery - -: easy of use websites, fairly priced tennis clothing . . I was also working and on a holiday, needing a next days delivery of, which No else other tennis store it looked can could confirm, when this package actually landed the same next day . Some. (58)
\end{tabular}

}
\caption{Rewriting examples (Trustpilot) from using Algorithm \ref{alg:rewrite2}. The numbers in parentheses denote the token length of the corresponding text. $A$ indicates the \textit{token addition probability}. A \textit{deletion probability} ($D$) of 0.05 was used in all examples. As shown, this new rewriting mechanism allows for output privatized texts of varying lengths.}
\label{tab:example4}
\end{table*}

\end{document}